\documentclass[conference]{IEEEtran}
\IEEEoverridecommandlockouts
\usepackage{cite}
\usepackage{amsmath,amssymb,amsfonts}
\usepackage{algorithmic}
\usepackage{graphicx}
\usepackage{textcomp}
\usepackage{xcolor}
\usepackage{hyperref}
\def\BibTeX{{\rm B\kern-.05em{\sc i\kern-.025em b}\kern-.08em
    T\kern-.1667em\lower.7ex\hbox{E}\kern-.125emX}}
\begin{document}

\title{Lessons Learned from Educating AI Engineers\\
\thanks{This work is financed by the Taskforce for Applied Research (SIA), part of the Netherlands Organisation for Scientific Research (NWO)}
}

\author{\IEEEauthorblockN{Petra Heck}
\IEEEauthorblockA{\textit{Fontys University of Applied Sciences} \\
Eindhoven, Netherlands \\
p.heck@fontys.nl}
\and
\IEEEauthorblockN{Gerard Schouten}
\IEEEauthorblockA{\textit{Fontys University of Applied Sciences} \\
Eindhoven, Netherlands \\
g.schouten@fontys.nl}
}

\maketitle

\begin{abstract}
Over the past three years we have built a practice-oriented, bachelor level, educational programme for software engineers to specialize as AI engineers. The experience with this programme and the practical assignments our students execute in industry has given us valuable insights on the profession of AI engineer. In this paper we discuss our programme and the lessons learned for industry and research.   
\end{abstract}


\section{Introduction}
Fontys University of Applied Sciences (Fontys UAS) is a higher education institution with profession-oriented study programs. One of the ICT (Information and Communication Technology) programmes at Fontys UAS focuses on Artificial Intelligence (AI). The goal of the ICT \& AI programme is to teach software engineering students how to engineer software systems that contain machine learning components. According to us this fits the definition of an AI engineer: a software engineer that builds an AI-based system. In the current state-of-the-practice at our applied university AI-based translates to "with machine learning techniques".

To educate our software engineering students as AI engineers we built a machine learning engineering programme on top of the software engineering programme.  We have built a programme around basic machine learning (including deep learning) that spans the entire machine learning life-cycle from business understanding until deployment. Our students perform assignments in or for industry in each semester, which provides us with feedback if the contents of the programme match what is needed in industry. This setup has given us a way to validate our understanding of the profession of AI engineer or machine learning (ML) engineer.

Since the start of the programme more and more has been published on AI engineering. Our current programme is partly based on literature review and partly on our experiences with industry assignments. We update our programme each semester based on the latest insights. The goal of this paper is to share our experience in keeping up education with such a fast evolving field.     

We start by summarizing the programme in terms of content and tools used. As an example we also show three of the industry projects that our students worked on. Then we discuss our programme by highlighting some of the lessons learned for AI engineers and linking it to MLOps. We end the paper with related work and our conclusion. 

\section{ICT \& AI programme at Fontys UAS}
\begin{figure}
\centering
\includegraphics[width=0.9\columnwidth]{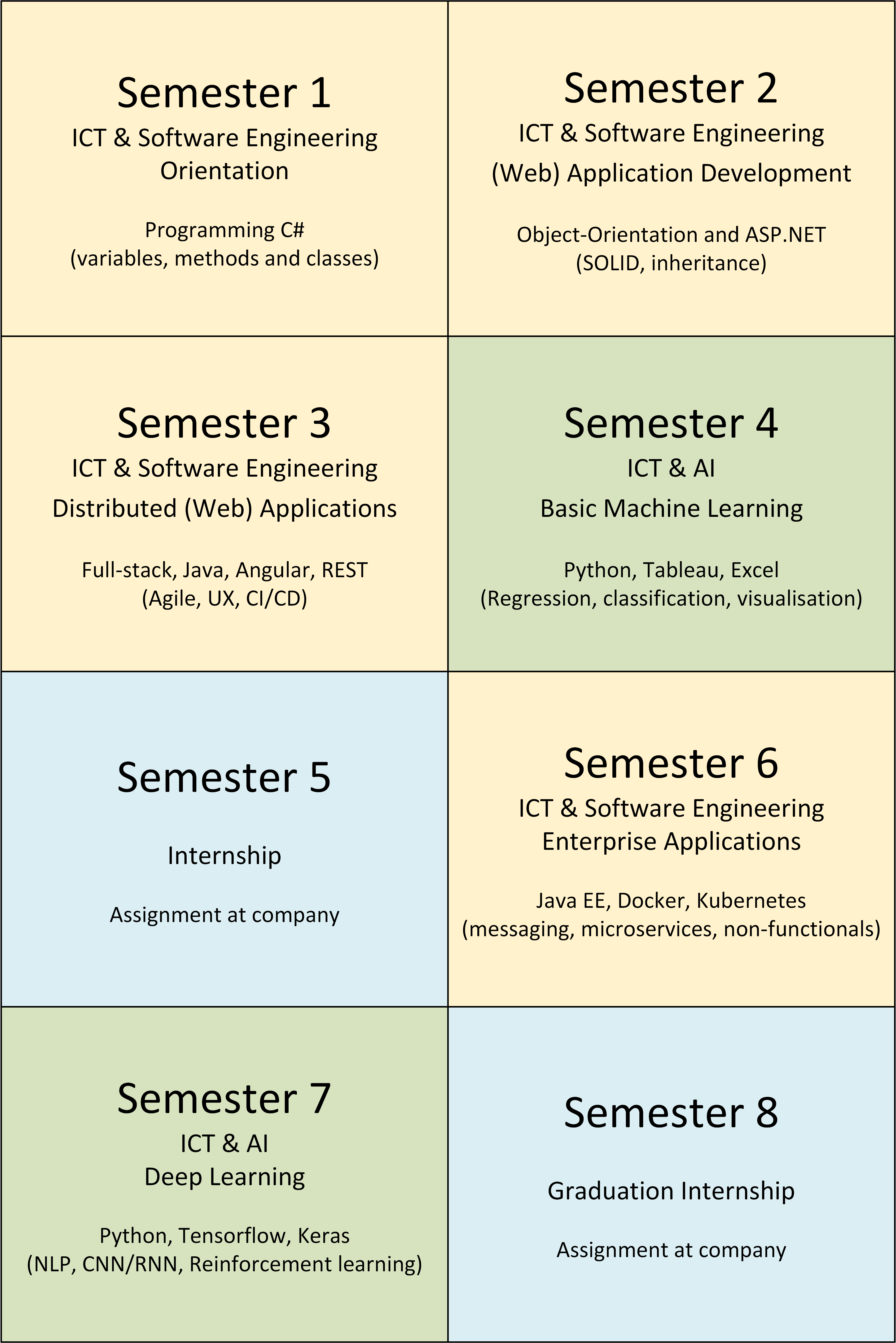}
\caption{Software Engineering \& AI programme at Fontys UAS}
\label{fig_curAI}
\end{figure}

ICT \& AI is a specialization available to the students since September 2017. They choose this specialization next to a main ICT profile such as Software Engineering. Such a combined programme (see Figure \ref{fig_curAI}) consists of: 
\begin{enumerate}
    \item four semesters of applied Software Engineering
    \item two semesters of applied AI 
    \item two internships engineering AI applications at and for an external organisation
\end{enumerate}
In June 2020, the first group of students that followed this complete programme graduated.

We realized that such a modular way of offering AI modules on top of Software Engineering modules also could work for software engineers in practice. They have already been educated as software engineers and thus only need to follow the AI modules. 
\subsection{Content of Programme}
As can be seen in Figure \ref{fig_curAI} semester 4 offers workshops on basic machine learning with Python. There are also workshops on data engineering (data quality, cloud storage) and data visualization. Furthermore we included workshops on the organizational and societal (ethics, law) impact of machine learning applications. 

In Semester 7 we extend the knowledge of machine learning algorithms by offering workshops on deep learning (CNN and RNN) and reinforcement learning. We also added workshops on Natural Language Processing to zoom in on the techniques needed to prepare textual data for the application of machine learning. In Semester 7 we reserve each Wednesday afternoon for inviting guest lecturers. This gives us the opportunity to highlight the state-of-the-practice not included in the base material of the semester yet and to enrich the theory with stories and examples from AI engineers in industry.

During both semesters the students do an individual and a group project where they complete the full cycle from collecting data from a client to reporting model results (and impact) to a client. Semester 7 has less workshops than Semester 4. The students are more self-directed to collect the necessary knowledge online, and they build upon the workshops they received in Semester 4. This is the same an AI engineer would do in practice and prepares them for their individual graduation projects in Semester 8. \\

\subsection{Example Projects}
As said, each semester is built around a case that comes from an external client (company or other organisation). This gives us a good view of real-world AI applications and the problems that arise when trying to apply AI methods in practice. The students do these projects in groups, mentored by teachers in the role of coach or subject matter expert. Semester 5 and 8 are internships where the student works independently at the external organization. Supervision is mostly done by a mentor from that external organization. In this section we will give some example projects that students worked on in the last three years.

\subsubsection{Reinforcement Learning}
Teach a robot arm how to learn the exact position of keyboard buttons on a virtual keyboard on a mobile device. Students applied reinforcement learning and built a simulator application to simulate the robot arm pressing keys on the virtual keyboard.

\subsubsection{Fraud Detection}
Detect fraudulent transactions for a tank card. Students used K-means clustering to detect outliers and built a complete data pipeline to process from raw transaction data to fraud predictions.  

\subsubsection{Matching}
Improve the matching of work shifts and healthcare professionals. The student implemented a complete data pipeline and deployed the trained Random Forest Classifier to a cloud environment. 

Each of the example projects clearly shows that a combination of software engineering skills and machine learning skills is needed to fulfill the assignment. For more complicated machine learning models our students might need to work with (master-level) data scientists. To work together with data scientists our students need to speak the language of data scientists. That is what they learn by doing simple machine learning models themselves.  

\subsection{Tools}
\begin{figure}
\centering
\includegraphics[width=0.9\columnwidth]{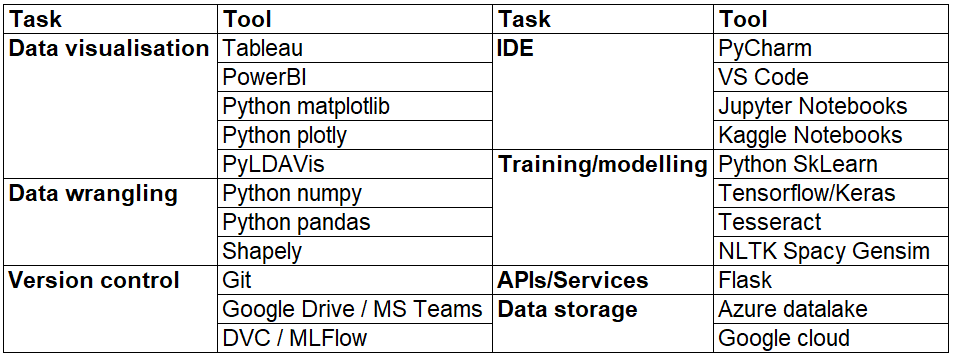}
\caption{Examples of tools and frameworks used at Fontys UAS}
\label{fig_tools}
\end{figure}

Students are free to use whatever tools or frameworks they deem necessary to fulfill their assignments. In the teacher material we use Python (scikit-learn, matplotlib, pandas), Tensorflow (Keras), Tableau, and Excel for data processing and machine learning. For two years in a row we interviewed student groups in Semester 7 to find out which other tools they use. Figure \ref{fig_tools} shows the list of tools and frameworks. In the second year we specifically introduced tools for version control in machine learning like MLFlow (mlflow.org) and DVC (dvc.org) because we saw that version control was not well managed in the first year (Git, Google Drive and Microsoft Teams). Students reported good experiences with the additional version control tooling to keep track of their data wrangling, feature selection and model building experiments. 

\section{Discussion}
In this section we will discuss the above programme,  resulting in lessons learned for AI engineers. 

\begin{figure}
\centering
\includegraphics[width=0.9\columnwidth]{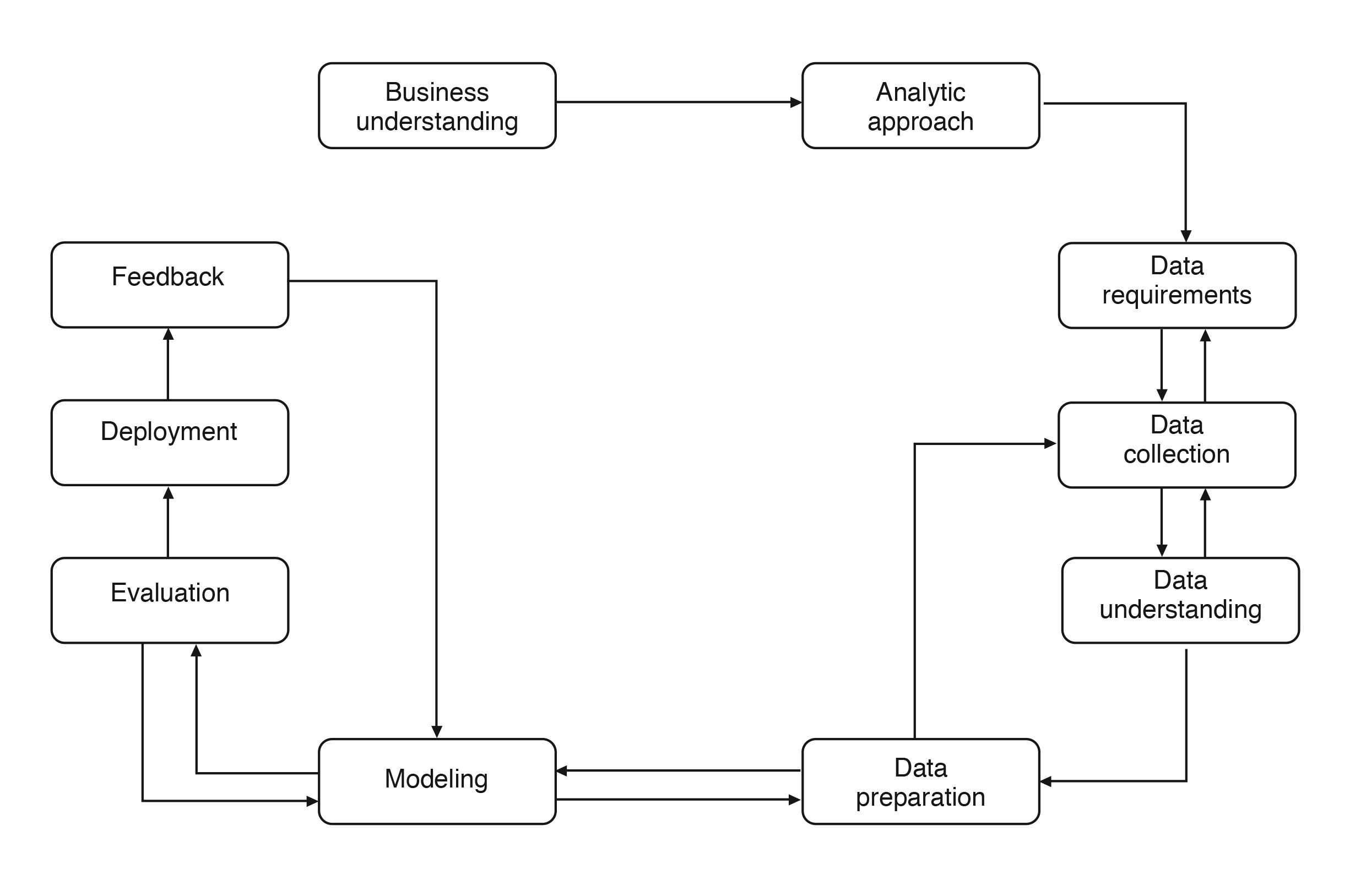}
\caption{IBM Data Science Methodology\cite{IBM}}
\label{fig_IBM}
\end{figure}

We teach our students to follow a structured approach in all their projects. For the ICT \& AI semester we selected the IBM Data Science Methodology \cite{IBM} (based on CRISP-DM\cite{CRISP}), see Figure \ref{fig_IBM}. The challenge lies in combining this data-driven process with the code-driven process of software engineering\cite{Heck}.   

\begin{center}\fbox{\parbox{8cm}{\textbf{Lesson Learned 1} For AI engineering the software development process needs to be extended with data and model engineering\cite{Heck}. We use the IBM Data Science Methodology\cite{IBM} with CRISP-DM\cite{CRISP} to understand the additional steps needed.}}\end{center}

As is clear from Figure \ref{fig_IBM} many steps in the machine learning workflow deal with data. In Semester 4 we spend quite a lot of time with the students to analyze, clean, and transform tabular data (with Excel or Python) to give them a grasp of the importance of the quality of the dataset for further analysis with machine learning models. We see that students are well-prepared for Semester 7, not so much because of the basic machine learning skills, but rather because they understand the importance of data engineering as a first step. 

\begin{center}\fbox{\parbox{8cm}{\textbf{Lesson Learned 2} An AI engineer spends a considerable amount of time on data engineering. Mastering data collection, data storage and data cleaning is a must.}}\end{center}

In the projects that the students do during Semester 4 and Semester 7, they have to communicate the results to external clients. These clients are not skilled in machine learning, so they do not know what terms like "accuracy", "training error", or "loss" mean. The students need another way of communicating results to the client. We found that the best way to do this is with plots. Data visualization is a skill that needs to be learned, as it is not present in the standard software engineering education. We teach the students types of plots and how to create them with tools like Python (matplotlib or plotly), Excel or Tableau. These types of visualizations have also been proven crucial in the "Data understanding" step (see Figure \ref{fig_IBM}). Plots help the students to get an overview on the data and spot areas to explore with again more detailed plots. These same plots can easily be shown to the client to discuss and confirm the students understanding of the data.   

\begin{center}\fbox{\parbox{8cm}{\textbf{Lesson Learned 3} An AI engineer needs to communicate with domain experts and clients on data and model performance. Mastering data visualisation is a must.}}\end{center}

Our students do not have a mathematical background (other then high school mathematics). Mathematics is not a subject in our ICT education. Because of this we had to explain all machine learning concepts on a conceptual level. Three years ago we were curious if students would be able to achieve valuable results without completely understanding the underlying mathematics. Today we know that the answer is yes. For our students, applying a machine learning algorithm, boils down to calling a function in Python. We need to teach them what types of parameters such functions have (and their conceptual meaning) and how to interpret the outcome. Furthermore we teach them a way of working in which validation is paramount, such that they know how to check the black-box results of the model. Our education does not provide students with the skills to change existing algorithms or build new ones, but we saw that valuable results for the clients can be achieved by applying existing algorithms or by re-using existing models. 

\begin{center}\fbox{\parbox{8cm}{\textbf{Lesson Learned 4} An AI engineer does not need to study advanced mathematics to become proficient in applying machine learning algorithms to solve practical problems.}}\end{center}

Our programme gives software engineers a grasp of what it takes to complete a machine learning project, such that they understand how to build software that contains a machine learning component. Advanced machine learning models need to be developed by master-level data scientists, but the expectation is that more and more algorithms and models become available for bachelor-level applied AI engineers to include in the software by themselves. Furthermore, the traditional data scientist needs the help of an AI engineer to incorporate the advanced machine learning model in a working software product. From our experience with graduation assignments we learned that our students, combining Software Engineering skills with basic machine learning skills, are perfectly capable of building software with a machine learning component.     

\begin{center}\fbox{\parbox{8cm}{\textbf{Lesson Learned 5} An AI engineer ought to combine knowledge on machine learning topics with knowledge on software engineering and DevOps: deployment of trained models, data pipelines, version control, testing, CI/CD, etc. An AI engineer needs to understand what is needed to deliver a production-ready software system that includes machine learning components.}}\end{center}

\begin{figure}
\includegraphics[width=\columnwidth]{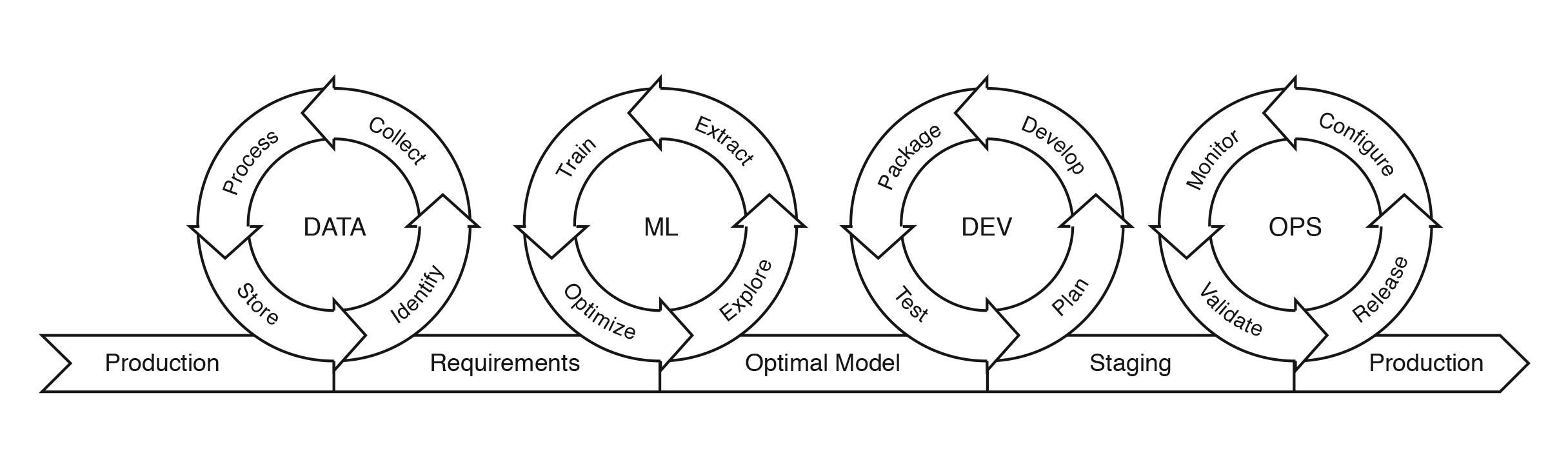}
\caption{MLOps\cite{Farah}}
\label{fig_MLOps}
\end{figure}

\section{MLOps and AI engineering}
We teach our students what it takes to build production-ready machine learning systems, based on what we know on building production-ready software systems. This extension of agile and DevOps to machine learning systems is also called MLOps\cite{Farah}, see Figure \ref{fig_MLOps}. Most publications on MLOps are written by practitioners (often grey literature such as blog posts) and thus are an excellent source for our practice-oriented students. Most publications on AI engineering are of academic nature and thus less applicable for our students. It is good to realize that AI engineering and MLOps have the same relation as software engineering and DevOps. This means that the research agenda in AI engineering should eventually lead to tools, frameworks and best practices to support practitioners to build production-ready machine learning systems following an MLOps approach. MLOps principles (\href{http://mlops.community/manifesto/}{mlops.community/manifesto/}) state that production-ready machine learning systems should:
\begin{itemize}
    \item be developed with a collaborative team across the full machine learning lifecycle from data to deployment;
    \item deliver reproducible and traceable results;
    \item be continuously monitored and improved.
\end{itemize}
The discipline of AI engineering should adhere to these principles as well. 

\begin{center}\fbox{\parbox{8cm}{\textbf{Lesson Learned 6} Researchers in AI engineering should collaborate with practitioners in MLOps to achieve a synergy that leads to the efficient and effective engineering of high-quality AI systems.}}\end{center}

\section{Related Work}
When designing our program we made use of the Edison Framework\cite{Edison}. The Edison Framework is a deliverable of an EU research project to establish a baseline for the upcoming profession of a data scientist. The Edison Framework identifies a number of profiles. Our programme most closely matches that of the (associate level) Data Science Engineer: develops large data analytics applications to support business processes. We used the Edison Framework to check for missing topics and knowledge areas and to frame our programme in relation to other data science programmes.

The course "Software Engineering for AI-enabled Systems (SE4AI)"\cite{SE4AI} has similar objectives to our programme, but is for master students. The master students work on a fictitious classroom project, whereas our students work on practical assignments from real companies. This gives our programme a more natural link to the profession of AI engineer. We did not find any other educational publications about AI engineering courses that we could compare to. 

Only recently research interest has been drawn to AI engineering\cite{Menzies}\cite{AI}. The last 2 years we have conducted a major literature review (including grey literature like blogs and workshop presentations) on this topic\cite{Heck}\cite{Heck2}\cite{Heck3}. This resulted in an update of the programme with much more emphasis on the software engineering side of AI engineering, as the first version of our programme was too much focused on the machine learning side. As said, our students do not have the mathematical background to develop new AI algorithms, so we needed to shift the programme to the skills needed to apply existing algorithms in software systems. 

In his book on machine learning engineering, Burkov\cite{Burkov} shares our vision: "Machine learning engineering (MLE) is the use of scientific principles, tools, and techniques of machine learning and traditional software engineering to design and build complex computing systems. MLE encompasses all stages from data collection, to model building, to making the model available for use by the product or the customers." Our educational programme provides a practical implementation of all steps included in the book, for students without a deep mathematical background.

The recently picked-up interest in SE for AI led to several research agendas, like the one presented by Bosch et al.\cite{Bosch2}. These research agendas are in fact also an agenda for the education of AI engineers.

\section{Conclusion}
After ample experience with educating software engineers we now also have over three years of experience in educating AI engineers. We share our experience through this paper.    
This paper contributes in the following ways:
\begin{enumerate}
    \item We have presented our bachelor-level ICT \& AI programme as a way to observe applied AI engineering. 
    \item We have presented methods, tools, techniques and frameworks that we use to educate our applied AI engineers. 
    \item We have summarised six lessons learned for applied AI engineers.
\end{enumerate}
The educational programme that we have is successful in educating applied AI engineers. We plan to verify if a similar programme could also work for practitioners switching from software engineer to AI engineer. The lessons we learned are also valuable for software organizations embarking on their own paths towards building AI applications and platforms. We will keep updating our body of knowledge to the newest developments in the fast evolving field of AI engineering.

\vspace{12pt}
\end{document}